%% file: acl_latex.tex
\pdfoutput=1

\documentclass[11pt]{article}

\usepackage[preprint]{acl}

\usepackage{times}
\usepackage{latexsym}

\usepackage[T1]{fontenc}

\usepackage[utf8]{inputenc}

\usepackage{microtype}

\usepackage{inconsolata}

\usepackage{graphicx}
\usepackage[inkscapelatex=false]{svg}

\usepackage{amsmath}
\usepackage{amssymb}

\usepackage{multirow}

\usepackage{makecell} 

\title{Beyond Independent Passages: Adaptive Passage Combination Retrieval for Retrieval Augmented Open-Domain Question Answering}

\author{
 \textbf{Ting-Wen Ko \textsuperscript{1,3} \thanks{Work done while in National Taiwan University.}},
 \textbf{Jyun-Yu Jiang\textsuperscript{2}},
 \textbf{Pu-Jen Cheng\textsuperscript{1}}
\\
 \textsuperscript{1}National Taiwan University,
 \textsuperscript{2}Amazon Search,
 \textsuperscript{3}University College London
}

\begin{document}
\maketitle
\begin{abstract}
Retrieval-augmented generation (RAG) enhances large language models (LLMs) by incorporating external documents at inference time, enabling up-to-date knowledge access without costly retraining. However, conventional RAG methods retrieve passages independently, often leading to redundant, noisy, or insufficiently diverse context—particularly problematic in noisy corpora and for multi-hop questions. To address this, we propose Adaptive Passage Combination Retrieval (AdaPCR), a novel framework for open-domain question answering with black-box LMs. AdaPCR explicitly models dependencies between passages by considering passage combinations as units for retrieval and reranking. It consists of a context-aware query reformulation using concatenated passages, and a reranking step trained with a predictive objective aligned with downstream answer likelihood. Crucially, AdaPCR adaptively selects the number of retrieved passages without additional stopping modules. Experiments across several QA benchmarks show that AdaPCR outperforms baselines, particularly in multi-hop reasoning, demonstrating the effectiveness of modeling inter-passage dependencies for improved retrieval.
\end{abstract}

\input{intro}

\input{related}
\input{methodology}
\input{experiment}
\input{result}
\input{conclusion}
\input{limitations}
\input{acknowledgements}

\bibliography{custom}

\appendix
\input{appendix}

\end{document}

%% file: intro.tex
\section{Introduction}

Recent advances have made large language models (LLMs) \citep{brown2020language, touvron2023llama} highly performant across a range of tasks. However, their internal knowledge becomes quickly outdated, and the large size makes fine-tuning on small, time-sensitive datasets costly. Retrieval-augmented generation (RAG) addresses these challenges by inserting external documents into the prompt, updating the model's knowledge without retraining. This simple yet powerful idea has made RAG an increasingly popular framework. \cite{gao2023retrieval} Consequently, a significant research focus has emerged: the application of RAG to frozen, black-box LLMs for downstream tasks, notably open-domain question answering (ODQA) \cite{zhang2024retrievalqa}.


Existing RAG approaches typically retrieve a fixed number of passages, either by appending them to the prompt \cite{ic-ralm} or by integrating them during decoding \cite{knnlm}. However, these approaches treat passages independently, overlooking the interplay between documents. This creates challenges in several scenarios. First, when dealing with noisy corpora—common in real-world web-scale datasets \cite{web-is-oyster, internet-aug-lm}—top-ranked documents are often highly redundant. Redundancy not only wastes valuable context windows but also crowds out potentially complementary information, leading to incomplete or suboptimal answers \cite{lost-in-middle}. 
Second, for multi-hop questions requiring reasoning over multiple sub-questions, retrieval must be contextually aware of previously selected documents; simple top-k retrieval based solely on the original query often fails to capture the necessary intermediate information. Third, the number of relevant documents can vary substantially: forcing a fixed number of retrieved passages may introduce irrelevant content, thereby distracting the LLM and degrading answer quality \cite{jin2024long, cuconasu2024power}.





In this work, we propose Adaptive Passage Combination Retrieval (AdaPCR), a retrieval framework that directly addresses these challenges for ODQA with black-box LMs. Instead of retrieving passages independently, AdaPCR jointly selects passage combinations, capturing document dependencies critical for complex and multi-hop questions.
AdaPCR operates in two stages: first, dense retrieval of candidate passages; second, a combination reranking stage where new queries are formed by concatenating initial passages with the question to refine retrieval based on context.
We train using a predictive reranking objective \cite{ic-ralm}, aligning retriever scores with downstream answer likelihood without fine-tuning the LLM. Additionally, we dynamically filter examples with no relevant documents, ensuring robust learning signals.
Our experiments demonstrate that AdaPCR significantly outperforms baselines across multiple QA benchmarks, particularly excelling in multi-hop reasoning tasks. 

We summarize our contribution as follows:
\begin{itemize}
    \item We propose AdaPCR, a novel retrieval framework for Open-Domain Question Answering with black-box language models that directly addresses the limitations of independent passage retrieval by explicitly modeling passage dependencies.
    \item We utilize a predictive reranking objective, aligning retriever scores with downstream answer likelihood from a black-box LLM without requiring its fine-tuning.
    \item We demonstrate through extensive experiments the performance of AdaPCR on several standard QA benchmarks, with particularly strong gains on complex multi-hop reasoning tasks.
\end{itemize}

%% file: related.tex
\section{Related Work}

\subsection{RAG with Language Models}
Integrating retrieval into language generation pipelines has been widely studied \cite{fan2024survey}. Early works like kNN-LM \cite{knnlm} interpolate next-token distributions between language models and k-nearest neighbor results from a retrieval corpus. While simple, such approaches underutilize LMs’ reasoning capabilities over retrieved documents. 
Other methods concatenate each query with a single retrieved document and integrate their outputs probabilistically \cite{rag-ki-nlp, replug, ATLAS, REALM}, or perform this integration at the encoder level \cite{izacard2021leveragingpassageretrievalgenerative}. More recent approaches append all retrieved documents to the input \cite{ic-ralm}, enabling the LM to semantically reason across multiple passages. Although this risks exceeding input length limits, techniques like token truncation mitigate it \cite{ic-ralm}. We follow this multi-passage input pipeline to fully leverage LLMs’ semantic understanding.

\subsection{Optimizing the Retriever for Black-Box Language Models}

To augment LMs with retrievers, some works use off-the-shelf retrievers like BM25 \cite{bm25} or DPR \cite{dpr} for their direct use and simplicity \cite{izacard2021leveragingpassageretrievalgenerative, schick2023toolformer}. 

However, these retrievers often misalign with the LM’s generation needs as they are designed for standalone retrieval tasks like search, instead of the augmentation in LM. Thus, many works propose fine-tuning retrievers using language modeling perplexity \cite{replug, ic-ralm} or answer likelihood \cite{ic-ralm, zhang2023retrieveaugmentlargelanguage} while keeping the LM frozen. Building on these ideas, our work improves retriever optimization by explicitly modeling document dependencies and introducing a new training objective with stabilized sampling, leading to enhanced open-domain QA performance.


Reranking techniques are employed to enhance retrieval quality for QA \cite{glass2022re2gretrievererankgenerate, ren2023rocketqav2jointtrainingmethod, iyer2021reconsider, wang2021retrieval, kruit2024retrieval}. These methods calculate a granular score representing the relationship between the query and each individual passage, utilizing these scores for subsequent re-ordering. Nevertheless, a limitation is that as each passage is scored independently, it overlooks the potentially important interactions between passages, which can be important for multi-hop QA or in noisy corpus scenarios. To address this by explicitly modeling relationships between documents, alternative retrieval architectures, such as tree-structured indices \cite{sarthi2024raptor} or graph-neural-network-enhanced retrievers \cite{li2024graphneuralnetworkenhanced} has been explored. Our AdaPCR similarly models these inter-passage relationships via its combination retrieval framework.





\subsection{LLM Guided Pre- and Post-Retrieval Techniques}
In addition to directly improving the retrievers and rerankers, some pre- and post-retrieval techniques are also used to improve the quality of the retrieval. 

Pre-retrieval techniques include query rewriting and expansion using LLMs \cite{xu2024search, jiang2025deepretrievalhackingrealsearch, su2022context, ircot, kim2024qpaug, sarthi2024raptor}, enabling broader and more context-aware document retrieval. On the other hand, post-retrieval techniques refine retrieved documents. For example, Re2G \cite{glass2022re2gretrievererankgenerate} combines and reranks dense and sparse retrieval results to improve generation quality. Some approaches use LM confidence to filter retrieved documents \cite{wang2023self, asai2024selfrag, yu2023improvinglanguagemodelsplugandplay}, but overreliance on LM confidence can misidentify noisy documents. Other works attempt to preemptively remove known knowledge via LM prompting \cite{wang2024blendfilteradvancingretrievalaugmentedlarge}, but these approaches risk inaccuracies due to unfaithful internal reasoning of LMs \cite{lanham2023measuring}.
In contrast, our method avoids document filtering during inference and instead directly optimizes the retriever distribution to align with answer likelihood, assigning lower weights to less useful documents without discarding them.

\subsection{Overcoming Noisy Retrieved Result}
Effective retrieval must address not only relevance but also diversity and redundancy. This is particularly critical when dealing with large, noisy corpora like web-scale datasets \cite{web-is-oyster}. To fix this problem in RAG, some approaches incorporate relevance thresholds and Maximal Marginal Relevance (MMR) \cite{yasunaga2022retrieval, vikraman2021passage, gao2024vrsd}, clustering-based selection \cite{xu2024tri}, or identifying diverse facets of the query before retrieval \cite{diverseargumentsearch}. Particularly, adaptive retrieval stopping \cite{kratzwald2018adaptive} predicts when sufficient information has been retrieved, to which ours is similar in having a dynamic selection of retrieved documents but without requiring a separate stopping classifier.

Rather than fixing retrieval alone, another strategy is to make LM robust against noisy inputs. Several works synthesize noisy contexts for LM training \cite{yoran2023making, lyu2023improving}, or use test-time adaptation to mitigate retrieval errors \cite{hübotter2025efficientlylearningtesttimeactive}. While these techniques complement our work, our focus remains on improving retriever quality to reduce noise at its source.

%% file: methodology.tex
\section{Methodology}
Our retriever framework addresses the challenge of identifying optimal passage combinations that maximize answer generation likelihood. Unlike traditional approaches that select passages independently, our method explicitly models dependencies between passages to improve question answering performance.

\subsection{Adaptive Passage Combination Retrieval (AdaPCR)} \label{sec:problem-formulation}

Given a question \(x\), a large set of passages \(\mathcal{C} = \{ p_i \}\), and a gold answer \(y\), our goal is to find a subset of passages \(\mathbf{d} \subset \mathcal{C}\) that enables a language model (LM) to accurately generate the answer \(y\) when conditioned on both \(x\) and \(\mathbf{d}\).

Prior methods typically assume that a fixed number \(k\) of passages are retrieved independently based on their relevance to \(x\). However, when answers require combining information across passages, retrieving passages independently can be suboptimal. Some passages may be individually unhelpful but crucial when considered together.

Thus, we aim to retrieve passage \emph{combinations} that maximize the downstream answer generation likelihood:
\[
\hat{\mathbf{d}} = \arg\max_{\mathbf{d}} P_{\text{LM}}(y \mid [\mathbf{d}; x]),
\]
where
\begin{equation}
    \begin{aligned}
        P_{\text{LM}}(y|[\mathbf{d}; x]) &= P_{\text{LM}}(y_1, y_2, \ldots, y_n|[\mathbf{d}; x]) \\
        &\triangleq \prod_{i=1}^{n} P_{\text{LM}}\left(y_{i} \mid [\mathbf{d}; x; y_{<i}]\right).
        \label{eq:lm_score}
    \end{aligned}
\end{equation}

We propose explicitly modeling passage dependencies by learning a \emph{combination retriever} \(P_{\text{ret}}(\mathbf{d} \mid x)\) that scores and selects groups of passages jointly rather than individually. We adopt the common bi-encoder \cite{dpr} structure as our retriever, including a query encoder $E_q$ and a passage encoder $E_d$.

Concretely, our framework proceeds in two stages:
\begin{enumerate}
  \item \textbf{First-Stage Retrieval:} We score all passages and retrieve a small pool of top-\(k\) candidate passages \(D_1=\{p_1, \dots, p_k\}\) based on their relevance to the question \(x\).
  
  \item \textbf{Combination Reranking:}  
  For each retrieved document $d_i$ from the first stage, we concatenated it with the input $x$ to form a new query $d_i \oplus x$. Then, we again score and choose the top $k$ passages relevant to the new concatenated query. This form a new passage combination candidate set \(D_2=\{p_{ij} | 1 \leq i, j \leq k\}\)  By so, we can score each combination \(\mathbf{d}\) using the same scoring model \(s(\mathbf{d}, x)\). Finally, the retriever selects the combination with the highest score, which can have one or two passages.
\end{enumerate}

We can then formally define the scoring function. In First-Stage Retrieval, we compute the cosine relevance scores of single passages $\mathbf{d} = \langle d_i \rangle$ formally as:
\begin{equation}
    s({\langle d_i \rangle}, x) \triangleq \cos(E_q(x), E_d(d_i)).
    \label{eq:ret_score}
\end{equation}

In Combination Reranking, we score a pair of passages $\mathbf{d}={\langle d_i, d_{ij} \rangle}$ as the relevance between the new query $d_i \oplus x$ and the second passage $d_{ij}$ as:
\begin{equation}
    s({\langle d_i, d_{ij} \rangle}, x) \triangleq  \cos(E_q(d_i \oplus x), E_d(d_{ij})).
    \label{eq:ret_score}
\end{equation}

The best scoring passage combination is then chosen by:
\begin{equation} 
d^* = \arg\max_{\mathbf{d} \in \mathcal{D}} s(\mathbf{d}, x),\quad D=D_1 \cup D_2, 
\end{equation} 

where $\mathcal{D}$ is the set of all candidate combinations, consisting of $k$ single-passage candidates $\langle d_i \rangle$ retrieved in the first stage, and 
$k \times k$ passage-combination candidates $\langle d_i, d_{ij} \rangle$ reranked in the second stage. 

This prevents redundancy in the combination as it rarely is the most helpful combination for answering when other passage combinations can provide more information to answer the question. By considering passage as part of the query, we also improve the system's understanding of the interplay between documents. This way, we implicitly balance between the relevance and diversity of the retrieved passages, finding the most informative and useful passage combination for question answering.

\subsection{Train the Reranker}
\label{sec:method-train-reranker}

To optimize our reranker, we adopt the predictive reranking approach \cite{ic-ralm}, which effectively aligns the retriever's preferences with the language model's performance, into our training objective. That is, we first normalize the retriever's confidence score:
\begin{equation}
P_{\text{ret}}(\mathbf{d} \mid x) = \frac{e^{s(\mathbf{d}, x) / \gamma}}{\sum_{\mathbf{d}' \in \mathcal{D}} e^{s(\mathbf{d}, x) / \gamma}},
\label{eq:P_ret}
\end{equation}
and then optimize the RAG-Sequence \cite{rag-ki-nlp} loss:
\begin{equation}
\mathcal{L} = -\sum_{i=1}^n \log P(y_i|x_i),
\end{equation}
where
\begin{equation}
P(y|x) \approx \sum_{\mathbf{d} \in \text{top-k}(P_{\text{ret}}(\cdot|x))} P_{\text{ret}}(\mathbf{d}|x) P_{\text{LM}}(y|[\mathbf{d}; x]).
\end{equation}

This formulation computes a weighted average of retriever confidence and language model answer score, providing a stable learning signal that naturally handles varying input ranges. Note that for different examples, $P_{\text{LM}}(y|[\mathbf{d}; x])$ are often of very different value ranges. Optimizing by this objective, examples with large $P_{\text{LM}}(y|[\mathbf{d}; x])$ thus naturally have relatively larger weights. This encourages the reranker to prioritize the examples with higher answer likelihood during optimization. Also, by keeping the wide range of answer likelihood values in the summation as-is, rather than normalizing it, this method has the advantage of numerical stability. We discuss more about this design choice in Section~\ref{sec:loss-discussion}.

\subsection{Training Data Sampling}

To train the retriever, we follow \cite{dpr} to use contrastive learning with in-batch negative sampling. However, we encounter a critical issue in training: the selection of positive training samples. Unlike an information retrieval training scheme where a golden passage exists for each example, in our case a question's answer can sometimes be absent in the top 100 passages. This can result in an ineffective learning signal, causing suboptimal convergence and performance during training. Therefore, we discard questions for which the correct answer is not found within the top 100 retrieved passages. This method significantly mitigates the initial training failure that arises when all selected documents are irrelevant to the question and the low probability of generating the correct answer.

%% file: experiment.tex
\section{Experimental Details}

\subsection{Datasets}
We evaluate our methods on two open-domain question-answering (ODQA) datasets: Natural Questions (NQ) \cite{naturalquestions} and HotpotQA \cite{hotpotqa}. NQ is representative of single-hop ODQA whereas HotpotQA is a multi-hop one. Following \citet{replug, ircot, izacard2021leveragingpassageretrievalgenerative}, we report Exact Match (EM) and Answer F1 score. 

\paragraph{Training split}
For each dataset, we randomly sample 10,000 data samples to form the training set. This sampling ensures that our models are trained on a representative subset of each dataset, facilitating effective learning and generalization. Additionally, for HotpotQA we additionally select 1,000 samples that are labeled as hard. This forms ``HotpotQA-hard'' training set to explore the improvement of retriever optimization when trained on a more context-aware dataset.

\paragraph{Testing split}
Retrievers trained on each dataset are evaluated using the corresponding ODQA test set. For NQ, we use 1,000 samples from the NQ evaluation set as our testing set. As for HotpotQA and HotpotQA-hard trained models, since the answers to the testing set are private, we follow the approach outlined by \cite{ircot} to randomly sample 1,000 data samples from the original development sets of HotpotQA as our evaluation set. We evaluate them in the one-shot setting.

\subsection{Model Configuration}
Our experiments include two components: a retriever and a generator. For retriever, we explore two dense retrievers:
\begin{itemize}
    \item \textbf{BERT} \cite{bert}: its pre-trained form, without additional fine-tuning on retrieval datasets. We aim to compare this with models like DPR, which are fine-tuned versions of BERT. This comparison helps to evaluate whether starting from a pre-trained BERT offers more room for improvement compared to starting from a fine-tuned model.
    \item \textbf{DPR} \cite{dpr}: we used two DPR variants in our experiments. For experiments on NQ, we used dpr-single-nq-base, which was a BERT model fine-tuned on the Natural Questions (NQ) dataset. For TriviaQA and HotpotQA, we used dpr-multiset, which is instead trained on NQ, TriviaQA, WebQuestions \cite{web-questions}, and TREC \citep{TREC-1, TREC-2}. These models are widely used in several RAG works \cite{rag-ki-nlp}.
\end{itemize}

For the frozen generator, we use Flan-T5-large. It has 780 million parameters on its encoder-decoder architecture and is designed to handle complex language tasks with high-quality answers \cite{flan-t5}.

\subsection{Retrieval Details}
We use MSMARCO-v2 \cite{msmarco} as the basis retrieval corpus, which consists of 138.4 million passages obtained from web documents crawled by Bing. This corpus is based on real user queries from Bing, providing a diverse and representative set of documents for retrieval tasks. Due to computational constraints, for each question in our dataset, we pre-retrieve the top 100 passages from MSMARCO-v2 as a subcorpus for our experiments. This is performed using BM25 \cite{bm25}, a well-known sparse probabilistic model that ranks documents based on their relevance to the query. We implement it using the Pyserini \cite{pyserini} library, which provides efficient and effective tools for information retrieval.

For the hyperparameter $k$ that decides the number of passages retrieved in the first stage, we choose $k=5$ to balance efficiency and performance. During inference time, a passage combination containing a maximum of two passages is picked.

\subsection{Baselines}
We compare two variants of In-Context-RALM (IC-RALM) \cite{ic-ralm} as our baseline, which is a widely adopted method in recent RAG studies. The vanilla IC-RALM retrieves passage using off-the-shelf retrievers. Since the questions in our dataset are simple questions with short lengths, we omit the retrieval stride design for IC-RALM and perform a single retrieval for each question. Note that this implementation is the same as the LM-Supervised Retrieval \cite{replug}. On the other hand, the IC-RALM reranker adopts a predictive reranking approach to train the reranker. This is similar to our approach where only first-round retrieval is performed and trained. The biggest difference in our method is that we model passage combinations explicitly during training for adaptive retrieval. 

We also report a baseline of ours using an off-the-shelf retriever, termed AdaPCR. This is in contrast to the AdaPCR Reranker, where the reranker is further trained to align downstream answer likelihood.

For all experiments, we set $k=5$. For the IC-RALM series, we use DPR \cite{dpr} as an initial retriever and pick the top two documents during inference time as a fair comparison.

%% file: result.tex
\section{Result}

Our experiments demonstrate the effectiveness of our proposed passage combination method for training retrievers, showing significant improvements over previous approaches. Table \ref{tab:results} summarizes our findings across different models, training methods, and loss functions.

\subsection{Comparison with Baseline}
\begin{table*}[t]
    \centering
    \caption{Performance comparison with IC-RALM baseline. For HotpotQA, we report the performance of retrievers trained on (normal) HotpotQA and HotpotQA-hard together in one grid, denoted {<normal> / <hard>}. The units of EM and F1 are \%. Retrievers here are initialized using the dpr series.}
    \label{tab:icralm-comparison}
    \begin{tabular}{|l|l|ccccc|}
        \hline
        Dataset & Metric & \makecell{No \\ Retrieval} & IC-RALM & AdaPCR & \makecell{IC-RALM \\ (Rerank)} & \makecell{AdaPCR \\ (Rerank)} \\
        \hline
        \multirow{2}{*}{NQ} & EM & 6.70 & 22.22 & 19.62 & 22.82 & \textbf{23.32} \\
         & F1 & 12.38 & 31.85 & 28.79 & 32.07 & \textbf{32.34} \\
        \hline
        \multirow{2}{*}{TriviaQA} & EM & 11.61 & 32.33 & 40.44 & 42.34 & \textbf{42.94} \\
         & F1 & 18.62 & 38.98 & 49.01 & 49.87 & \textbf{50.35} \\
        \hline
        \multirow{2}{*}{HotpotQA} & EM & 12.21 & 12.81 & 17.11 & 18.92 / 19.52 & 19.52 / \textbf{19.92} \\
         & F1 & 18.71 & 17.94 & 24.39 & 26.49 / 26.97 & 27.40 / \textbf{27.74} \\
        \hline
    \end{tabular}
\end{table*}

Table~\ref{tab:icralm-comparison} presents the performance comparison between our method and IC-RALM baselines. Overall, our AdaPCR-based models consistently outperform IC-RALM variants across all datasets after reranker training, particularly excelling in multi-hop settings.

On NQ, AdaPCR Reranker achieves a 1.10-point improvement in EM and a 0.49-point improvement in F1 over IC-RALM Reranker. Although our off-the-shelf retriever (AdaPCR) initially slightly underperforms IC-RALM, reranker training enables it to surpass IC-RALM's performance.

On TriviaQA, AdaPCR already shows strong gains without fine-tuning, outperforming IC-RALM by 8.11 EM and 10.03 F1 points. After reranker training, AdaPCR Reranker further improves, achieving the highest EM (42.94) and F1 (50.35).

On HotpotQA, which involves more complex multi-hop reasoning, our method demonstrates the most substantial advantages. Even before reranker training, AdaPCR outperforms IC-RALM by 4.30 EM and 6.45 F1 points. After training, AdaPCR Reranker improves further, surpassing IC-RALM Reranker by 0.40–0.60 points in EM and 0.43–0.77 points in F1, depending on the HotpotQA variant (normal or hard).

These results highlight several key findings. First, although our pretrained retriever may initially trail IC-RALM on simpler tasks like NQ, it surpasses it after reranker training. Second, AdaPCR’s stronger performance on TriviaQA and HotpotQA indicates its effectiveness at handling both single-hop and multi-hop questions, with particularly strong gains in complex, reasoning-intensive settings. Therefore, the flexibility of our pipeline—allowing for dynamic passage selection and alignment with downstream objectives—contributes to its robustness across different QA tasks.


\subsection{Effect of Positive Sampling on Training Stability}

\begin{table}
    \centering
    \begin{tabular}{|c|c|cc|}
        \hline
        Dataset & Metric & Original & Positive\\
        \hline
        NQ & EM & 20.72 & 23.32 \\
         & F1 & 30.16 & 32.34 \\
        \hline
        TriviaQA & EM & 40.44 & 42.94\\
         & F1 & 49.01 & 50.35\\
        \hline
        HotpotQA & EM & 17.11 & 19.52 \\
         & F1 & 24.39 & 27.4\\
        \hline
    \end{tabular}
    \caption{EM and f1 with positive sampling.}
    \label{tab:positive-sampling}
\end{table}

Our balanced sampling strategy, which ensures the representation of both single-passage and two-round combinations during training, significantly improves training stability and convergence. As shown in Figure~\ref{tab:positive-sampling}, models trained with our positive sampling approach perform better than those trained with standard random sampling.
This improvement is particularly pronounced for multi-hop questions, where identifying complementary passage combinations is essential. The balanced sampling approach ensures the model consistently encounters informative passage pairs during training, facilitating the learning of dependencies between passages.


\begin{table*}[t]
    \centering
    \caption{Exact match performance comparison of different models and training methods. The highlighted ones are the most performant setting in each row. In our main experiments, RAG loss is thus utilized for its better performance.}
    \label{tab:results}
    \begin{tabular}{|l|l|cc|cc|cc|}
        \hline
        \multirow{2}{*}{Model} & \multirow{2}{*}{Train Data} & \multicolumn{2}{c|}{KL divergence} & \multicolumn{2}{c|}{CE loss} & \multicolumn{2}{c|}{RAG loss} \\
         &  & 1-round & 2 round & 1 round & 2 round & 1 round & 2 round \\
        \hline
        \multirow{5}{*}{DPR} & NQ & 21.52 & 23.22 & 21.92 & 22.52 & 22.82 & \textbf{23.32} \\
         & TriviaQA & 41.43 & 38.34 & 41.44 & 41.74 & 42.64 & \textbf{42.94} \\
         & HotpotQA & 18.52 & \textbf{19.92} & 18.42 & 18.72 & 18.92 & 19.52 \\
         & HotpotQA-hard & 18.52 & 19.42 & 18.52 & 19.02 & 19.52 & \textbf{19.92} \\
         & average & 25.00 & 25.23 & 25.08 & 25.50 & 25.98 & \textbf{26.43} \\
        \hline
        \multirow{5}{*}{BERT} & NQ & 9.11 & \textbf{13.41} & 10.71 & 11.21 & 12.31 & 11.71 \\
         & TriviaQA & 25.13 & 26.23 & 25.03 & 26.53 & 26.93 & \textbf{29.33} \\
         & HotpotQA & 12.81 & 13.91 & 13.91 & 14.72 & \textbf{16.02} & 15.12 \\
         & HotpotQA-hard & 13.11 & 13.61 & 15.52 & 13.91 & 14.82 & \textbf{15.92} \\
         & average & 15.04 & 16.79 & 16.29 & 16.59 & 17.52 & \textbf{18.02} \\
        \hline
    \end{tabular}%
\end{table*}

\subsection{Comparison of Loss Functions}
\label{sec:loss-discussion}
We explore three choices of loss function for training the retriever: KL divergence loss, Cross-Entropy (CE) loss, and RAG loss.

\paragraph{KL Divergence Loss} 
\citet{replug} uses KL loss to train the retriever with a frozen generator, where the divergence between normalized retriever scores and normalized LM answer probabilities is minimized. However, it suffers from sensitivity to input range due to softmax normalization. Specifically, different answer lengths lead to wide variations in answer likelihoods across examples. It is thus challenging to tune the temperature parameter ($\beta$), causing distributions to either collapse or become overly uniform. This instability limits the retriever's ability to differentiate fine-grained passage quality in the ODQA setting.

\paragraph{Cross-Entropy (CE) Loss} 
To mitigate issues of softmax normalization, we also explore CE loss, treating the best passage combination as the gold label and applying standard supervised learning. CE loss improves training stability but still treats passage combinations independently without fully leveraging the probabilistic nature of LM outputs.

\paragraph{RAG Loss}
To overcome these limitations, we eventually adopt the RAG loss described in Section~\ref{sec:method-train-reranker}. It models the answer probability as a weighted sum over multiple passage combinations, without requiring softmax normalization on the LM outputs. This approach naturally accommodates varying answer scores, avoids brittle softmax behavior, and provides a more stable training signal. 

Interestingly, however, we theoretically show that the optimal behavior of the retriever under RAG loss should have the same result as CE loss (Appendix~\ref{sec:convergence}). However, we don't have the budget to run the training until full convergence, so we leave the empirical validation of this finding to future work.

Table~\ref{tab:results} summarizes the exact match (EM) results across different datasets and training settings.
Among the three loss functions evaluated, RAG loss consistently outperforms CE loss, which in turn outperforms KL divergence loss.

In one-round training, the average improvement of RAG loss over CE loss is 0.9 points for DPR and 1.23 points for BERT.
In two-round settings, RAG shows even stronger gains, achieving up to 0.93 points improvement for DPR and 1.43 points for BERT.

These results indicate that RAG loss is particularly effective at capturing relevance relationships between passage pairs, highlighting the benefit of treating passage combinations in a more probabilistic and holistic manner.

Finally, the superior performance of RAG and CE losses over KL divergence loss underscores the limitations of the softmax approach and the advantages of our design choice.


\subsection{BERT vs DPR Performance}

While DPR outperforms BERT overall, our results show that BERT can still serve as an effective initial model checkpoint for passage combination retriever, even without specific fine-tuning for retrieval tasks (Table~\ref{tab:results}). This demonstrates that our method can substantially enhance the retrieval capabilities of general-purpose language models like BERT, making them viable options for retrieval tasks without extensive pre-training on large-scale retrieval datasets.


\subsection{Case Study: Combining Passages for Improved Retrieval} \label{sec:case-study-2}

\paragraph{Complementary Information Retrieval} We show a case study comparing IC-RALM versus our method in Figure~\ref{fig:case-study-1}. Consider the question: "Which band is from England, Fireflight or Dirty Pretty Things?" (true answer: Dirty Pretty Things). We observed clear differences between the methods. IC-RALM retrieved a first passage that matched keywords but lacked semantic relevance, and a second passage that redundantly discussed "pretty thing," providing no new information. In contrast, our method retrieved complementary evidence: the first passage confirmed facts about "Dirty Pretty Things," while the second established that "Fireflight" is an American band. This provided clear, non-overlapping clues supporting the correct answer.

Our approach excels by gathering diverse hints across passages, minimizing redundancy, and effectively combining information about both entities to support reasoning toward the correct answer.

\paragraph{Multi-hop Reasoning}
We further illustrate the strengths of our approach with a multi-hop reasoning example in Figure~\ref{fig:case2}, following the same setup.

The question asked: "The Argentine National Anthem was adopted 3 years after which event that led to the removal of Viceroy Baltasar Hidalgo de Cisneros?" (correct answer: May Revolution).

DPR retrieved two passages, but neither sufficed: Passage 1 focused on the Colombian national anthem—showing strong keyword overlap but no semantic relevance to Argentina or the historical event. Passage 2 mentioned the removal of Viceroy Cisneros and Argentina, making it somewhat relevant, but it failed to specify the key event (the May Revolution).
Thus, the model judged the question unanswerable and produced an incorrect response, illustrating the retriever’s difficulty in assembling and reasoning over partial information.

In contrast, our framework—using a trained BERT retriever with RAG loss—demonstrated robust multi-passage reasoning. Passage 1 discussed Cisneros' appointment, and Passage 2 explicitly mentioned the May Revolution and its connection to Cisneros' removal. Crucially, our retriever leveraged information from Passage 1 to reformulate the query to contain Viceroy of Redo, leading to the retrieval of Passage 2. This showcases our model’s ability to model dependencies across passages and dynamically extend the search space.


By synthesizing evidence from both passages, our method correctly identified the May Revolution as the answer. This highlights the strength of our framework in handling complex, multi-hop questions by retrieving and integrating complementary information from multiple sources, ultimately enabling more accurate and complete reasoning.

%% file: conclusion.tex
\section{Conclusion}
We introduce AdaPCR, a retrieval framework that adaptively selects and scores passage combinations to improve open-domain QA with black-box LMs. By modeling dependencies between passages and aligning retriever scores with answer likelihood, AdaPCR overcomes key limitations of traditional RAG approaches—handling redundancy, promoting diversity, and supporting multi-hop reasoning. Experiments across multiple QA benchmarks show that AdaPCR outperforms IC-RALM baselines, especially on complex questions, validating the benefits of passage combination modeling and adaptive retrieval.

%% file: limitations.tex
\section*{Limitations}

We explore and compare our methods to other methods that also  train retriever in a black-box generator setting. However, many work also explore concatenating answer probabilities produced by incorporating different passages in parallel or inference multiple times \cite{yoran2023answering, jiapeng2024treereviewstreebaseddynamic}. Also, while the current procedure involves two rounds of retrieval, this iterative process could be extended to more rounds to explore even more complex passage combinations. Some strategies for making LM robust against noisy inputs can also be supplemented with our method. We leave these extension for future work.



%% file: acknowledgements.tex



%% file: appendix.tex
\section{Case Study}
\label{sec:appendix}
\begin{figure*}[t] 
    \centering
    \includegraphics[width=0.9\textwidth]{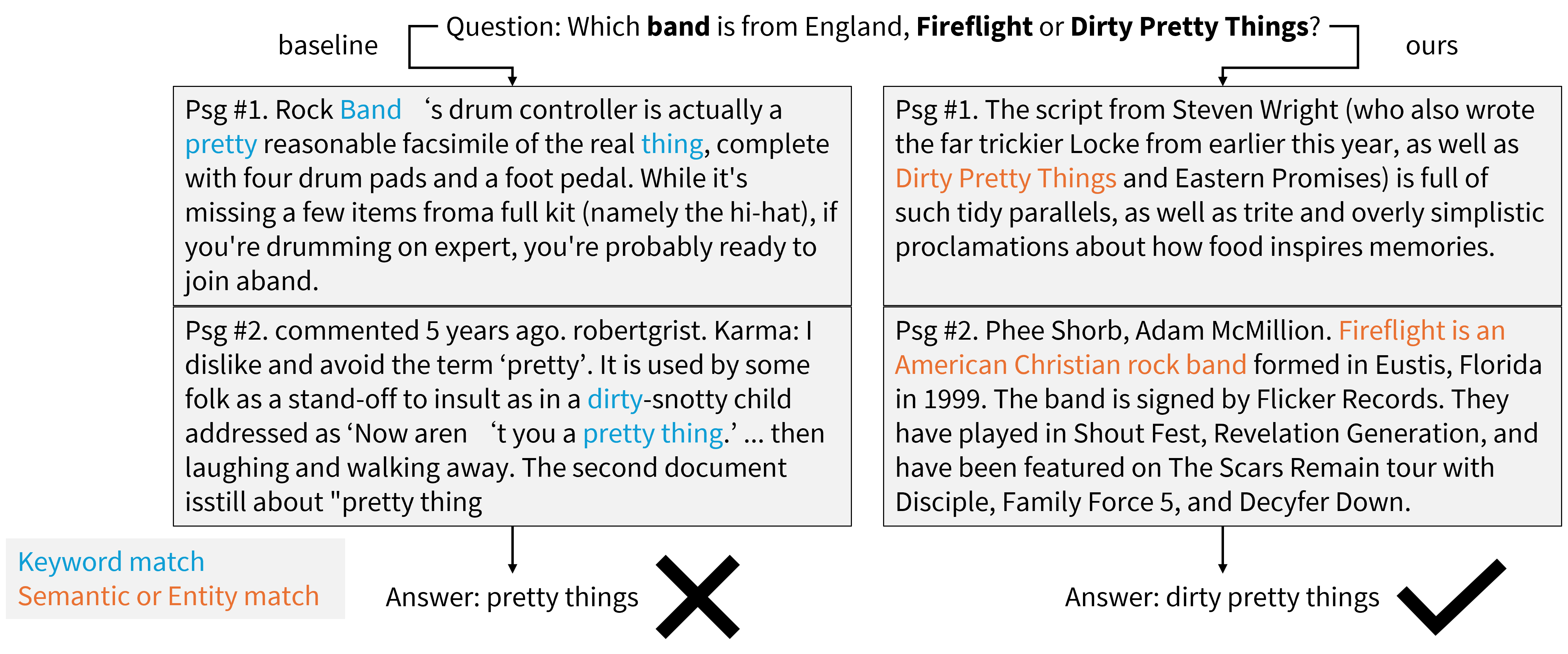}
    \caption{Case study 1: baseline vs. our method when asked the same question "Which band is from England, Fireflight or Dirty Pretty Things?" Note that the correct answer is "dirty pretty things"}
    \label{fig:case-study-1}
\end{figure*}

\begin{figure*}[t] 
    \centering
    \includegraphics[width=0.8\textwidth]{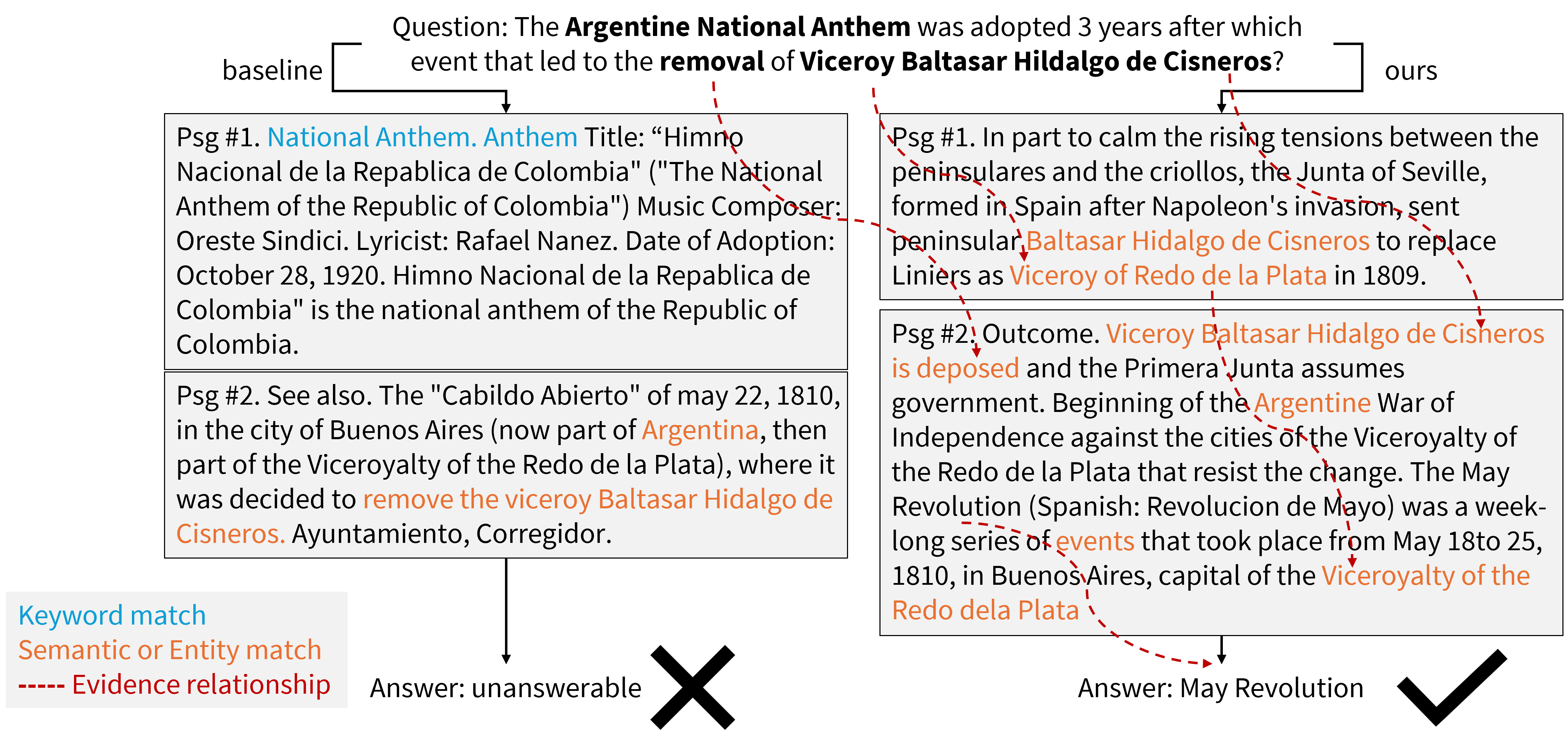}
    \caption{Case Study 2}
    \label{fig:case2}
\end{figure*}

Fig.~\ref{fig:case-study-1} and Fig.~\ref{fig:case2} demonstrates two case study of our method.

\section{Theoretical Analysis of Optimal Retriever Behavior}\label{sec:convergence}

In this section, we aim to demonstrate that the optimal retriever, which maximizes the probability of generating the correct answer \( P(y \mid x) \), is a one-hot retriever. This means that the optimal retriever selects the combination of passages that can bring the highest answer generation likelihood given the input \( x \). 

\paragraph{Setups}
First, by the normalization property of the retriever's output probabilities,
\begin{equation}
\sum_{i=1}^{n} P_{\text{ret}} \left( \mathbf{d}_i \mid x \right) = 1.
\end{equation}
Then, without loss of generality, we arrange the passage candidates in ascending order of their answer probabilities. That is, $\forall \, i \in [1, n-1]$,
\begin{equation}
P_{\text{LM}}\left(y \mid \left[\mathbf{d}_i; x\right]\right) \leq  P_{\text{LM}}\left(y \mid \left[\mathbf{d}_{i+1}; x\right]\right).
\end{equation}
In this setup, $\mathbf{d}_n$ is the best candidate passage that leads to the highest answer probability, which we aim to find in our optimization process of the retriever:
\begin{equation}
\mathbf{d}_n = \arg\max_{\mathbf{d}} P_{\text{LM}}\left(y \mid \left[\mathbf{d}; x\right]\right)
\end{equation}

\paragraph{Lemma 1}
The upper bound of the answer probability \textit{when given input \( x \)} is the maximum answer probability \textit{when given the best passage \( \mathbf{d}_n \) and input \( x \)}:
\begin{equation}
P_{\text{LM}}(y \mid x) \leq P_{\text{LM}} \left(y \mid \left[\mathbf{d}_n; x\right]\right)
\end{equation}

\paragraph{Proof of Lemma 1}
\begin{align*}
P_{\text{LM}}(y \mid x) &= \sum_{i=1}^{n} P_{\text{ret}} \left( \mathbf{d}_i \mid x \right) P_{\text{LM}} \left( y \mid \left[\mathbf{d}_i; x\right] \right) \\
&\leq \sum_{i=1}^{n} P_{\text{ret}} \left( \mathbf{d}_i \mid x \right) P_{\text{LM}} \left( y \mid \left[\mathbf{d}_n; x\right] \right)  \\
&= P_{\text{LM}} \left( y \mid \left[\mathbf{d}_n; x\right] \right) \sum_{i=1}^{n} P_{\text{ret}} \left( \mathbf{d}_i \mid x \right)  \\
&= P_{\text{LM}} \left( y \mid \left[\mathbf{d}_n; x\right] \right) \cdot 1  \\
&= P_{\text{LM}} \left( y \mid \left[\mathbf{d}_n; x\right] \right)
\end{align*}


\paragraph{Proposition 1} The one-hot retriever is an optimal retriever. Consider 
\[
\hat{P}_{\text{ret}} \left( \mathbf{d}_i \mid x \right) =
\begin{cases} 
1, & \text{if } i = n \\
0, & \text{otherwise}
\end{cases}
\]
then 
\begin{align*}
\hat{P}_{\text{ret}} &= \arg\max_{P_\text{ret}} P_{\text{LM}}(y \mid x) \\
&=\arg\max_{P_\text{ret}} \sum_{i=1}^{n} P_{\text{ret}} \left( \mathbf{d}_i \mid x \right) P_{\text{LM}} \left( y \mid \left[\mathbf{d}_i; x\right] \right)
\end{align*}

That is, the one-hot retriever is a optimal retriever in finding passages that maximize answer probability. This aligns our approach - optimizing retriever - with our eventual purpose - maximizing downstream answer probability:
\[
\arg\max_{\mathbf{d}} P_{\text{ret}} \left( \mathbf{d} \mid x \right) = \arg\max_{\mathbf{d}} P_{\text{LM}} \left( y \mid \left[\mathbf{d}; x\right] \right)
\]
\paragraph{Proof of Proposition 1}
When 
\[
\hat{P}_{\text{ret}} \left( \mathbf{d}_i \mid x \right) = 
\begin{cases} 
1, & i = n \\
0, & \text{otherwise}
\end{cases} \quad ,
\]

\begin{align*}
P(y \mid x) &= \sum_{i=1}^{n} \hat{P}_{\text{ret}} \left( \mathbf{d}_i \mid x \right) P_{\text{LM}} \left( y \mid \left[\mathbf{d}_i; x\right] \right) \\
&= 1 \cdot P_{\text{LM}} \left( y \mid \left[\mathbf{d}_n; x\right] \right) \\
&+ \sum_{i \ne n} 0 \cdot P_{\text{LM}} \left( y \mid \left[\mathbf{d}_i; x\right] \right) \\
&= P_{\text{LM}} \left( y \mid \left[\mathbf{d}_n; x\right] \right)
\end{align*}

From Lemma 1, we know:
\begin{align*}
P(y \mid x) &= \sum_{i=1}^{n} P_{\text{ret}} \left( \mathbf{d}_i \mid x \right) P_{\text{LM}} \left( y \mid \left[\mathbf{d}_i; x\right] \right) \\
&\leq P_{\text{LM}} \left( y \mid \left[\mathbf{d}_n; x\right] \right)
\end{align*}

Therefore,
\[
\hat{P}_{\text{ret}} = \arg\max_{P_\text{ret}} P_{\text{LM}}(y \mid x)
\]
